\pdfoutput=1
\documentclass[11pt]{article}

\usepackage[final]{acl}

\usepackage{times}
\usepackage{latexsym}
\usepackage{amssymb}
\usepackage{svg}
\usepackage{enumitem}
\usepackage{graphicx}
\usepackage{subcaption} % Modern replacement for subfigure

\usepackage{booktabs}

\usepackage[T1]{fontenc}
\usepackage[utf8]{inputenc}

\usepackage{microtype}

\usepackage{inconsolata}

\usepackage{graphicx}

\usepackage{xcolor}
\usepackage{mathtools}
\DeclareMathOperator*{\argmax}{argmax}

\title{ResFormer: All-Time Reservoir Memory \\for Long Sequence Classification}
\author{
  Hongbo Liu\textsuperscript{*} \quad Jia Xu \thanks{Authors are listed in alphabetical order.} \\
  Gateway Academic Center, Stevens Institute of Technology \\
  601 Hudson St, Hoboken, NJ 07030 \\
  \texttt{hliu93@stevens.edu}, \quad \texttt{jxu70@stevens.edu}
}

\begin{document}
\maketitle

%architecture figure udpate
%appendix
%compile errors
%supplementary material

\begin{abstract}
Sequence classification is essential in NLP for understanding and categorizing language patterns in tasks like sentiment analysis, intent detection, and topic classification. Transformer-based models, despite achieving state-of-the-art performance, have inherent limitations due to quadratic time and memory complexity, restricting their input length. Although extensive efforts have aimed at reducing computational demands, processing extensive contexts remains challenging. 

To overcome these limitations, we propose ResFormer, a novel neural network architecture designed to model varying context lengths efficiently through a cascaded methodology. ResFormer integrates an reservoir computing network featuring a nonlinear readout to effectively capture long-term contextual dependencies in linear time. Concurrently, short-term dependencies within sentences are modeled using a conventional Transformer architecture with fixed-length inputs. 

Experiments demonstrate that ResFormer significantly outperforms baseline models of DeepSeek-Qwen and ModernBERT, delivering an accuracy improvement of up to +22.3\% on the EmoryNLP dataset and consistent gains on MultiWOZ, MELD, and IEMOCAP. In addition, ResFormer exhibits reduced memory consumption, underscoring its effectiveness and efficiency in modeling extensive contextual information.

\end{abstract}

\section{Introduction}

Transformer models have significantly advanced state-of-the-art performance across various natural language processing (NLP)  tasks~\cite{vaswani2017attention,devlin2018bert,dosovitskiy2020image}. However, an important limitation of Transformer architectures is their inherent quadratic complexity concerning input length, which restricts the number of tokens they can effectively process. For instance, popular models such as LlaMA 3~\cite{llama3}, Gemma~\cite{team2024gemma}, GPT-4~\cite{achiam2023gpt}, and Mistral~\cite{jiang2023mistral} typically have a maximum input length limited to approximately $8K$ tokens. 

Despite this limitation, tasks such as emotion detection greatly benefit from analyzing extensive sequential contexts to capture subtle emotional cues and evolving relational dynamics effectively. For example, in the TV series \textit{Friends}, accurately understanding Joey's emotional development, from carefree bachelorhood to genuinely caring about his friends' happiness, necessitates context derived from numerous episodes spanning several seasons; Similarly, Monica’s compulsive cleanliness, initially played for laughs, but across seasons, it reflects deep insecurity rooted in childhood neglect. Without long-term context, such emotional complexity is easily misclassified. Thus, efficiently modeling all context of the narrative is essential to adequately analyze emotion.

Many studies have investigated how to efficiently increase Transformer input lengths, such as~\cite{munkhdalai2024leave,tworkowski2024focused,bertsch2024unlimiformer, mohtashami2024random}. 
Existing solutions, however, either modifying the attention model with heuristic assumptions or projecting long input into a fixed dimension. Since most work does not consider temporal patterns of the input, there is still room to reduce the information loss learned from the long context and improve the prediction accuracy and efficiency. 

 In this work, we introduce a novel neural network architecture ResFormer that integrates Reservoir Computing Network and Transformer to efficiently handle long sequences.  
Echo State Network (Reservoir) is a kind of Reservoir Computing (RC) that is a class of simple and efficient Recurrent Neural Networks where internal weights are fixed at random, and only a linear output layer is trained.
Reservoir requires a small number of training data samples and computing resources with  great advantages for processing sequential data in linear time and constant space~\cite{gauthier2021next}. Here, we improve Reservoir with nonlinear readout to take long conversational context into account for time and memory efficiency~\cite{gauthier2021next}. 

Notebly, our method is model agnostic, that means, our ResFormer can be also applied on any other neural network architecture such as convolutional neural network and LSTM to improve their performance. 

\begin{figure*}[!t]
  \centering
  \hspace{-1cm}
  %\includesvg[width=0.8\linewidth]{Figures/Figure1}
  \includegraphics[width=0.9\linewidth]{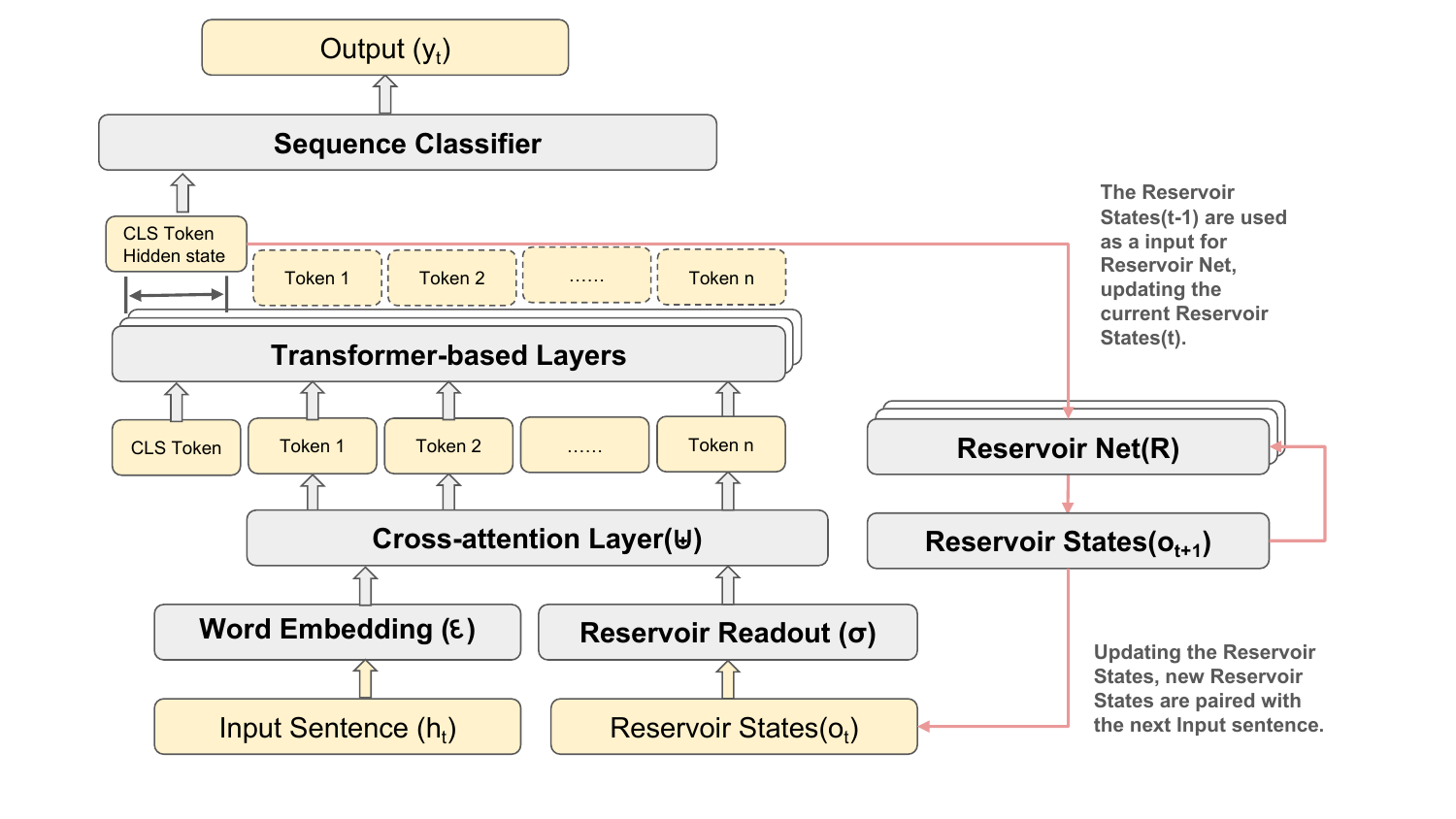}
    \caption{\small A schematic of ResFormer: input sentences are first processed by a word embedding layer, while Reservoir states are passed into a nonlinear readout layer. The resulting Reservoir outputs and input embeddings are then combined via a cross-attention layer and fed into the Transformer.}

  \label{fig:reservoir_transformer}
  \vspace{-1.2em}
\end{figure*}

\begin{figure}[!t]
  \hspace{-1cm}
  \includegraphics[width=1.2\linewidth]{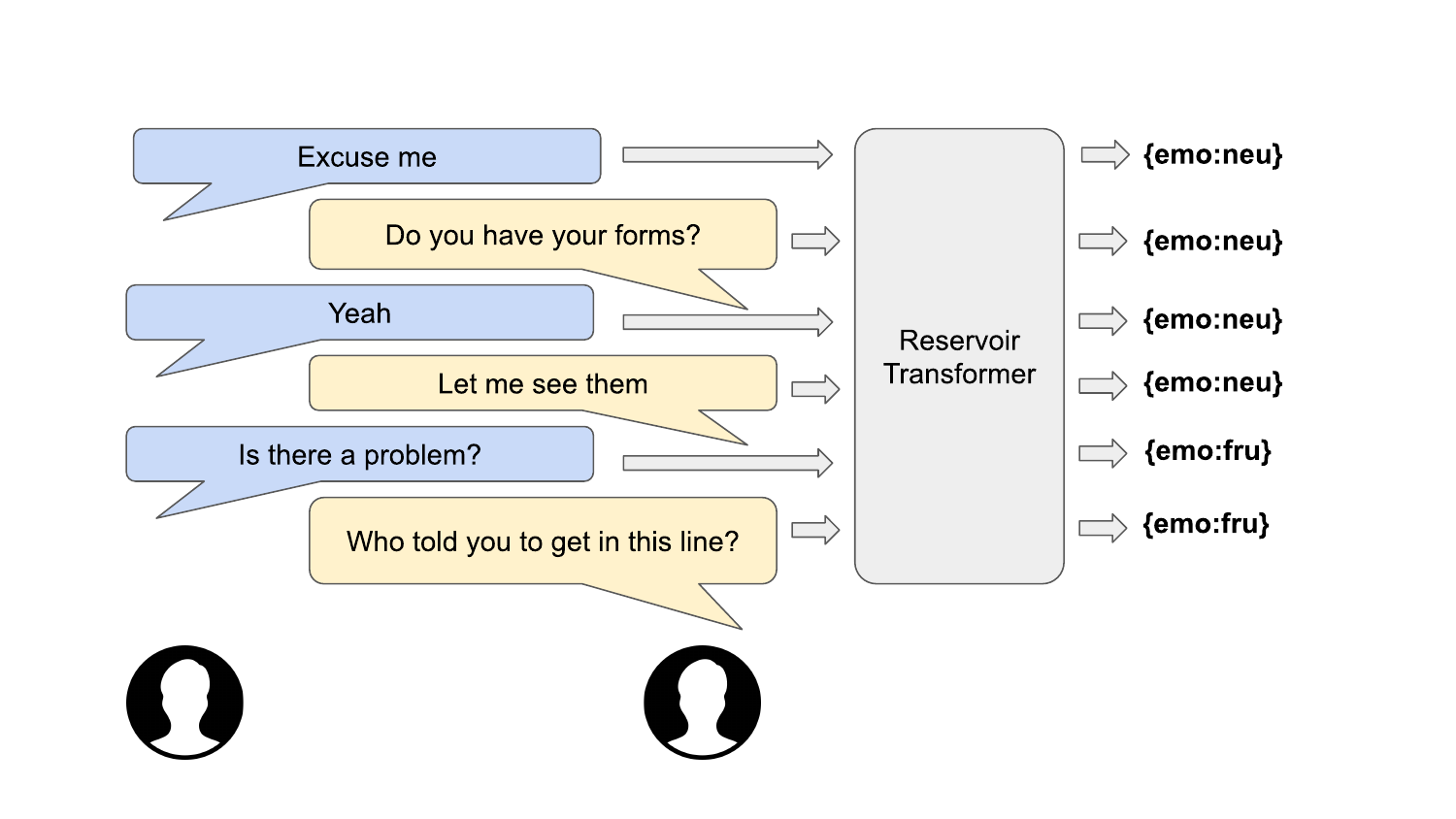}

    \caption{\small An example from the IEMOCAP dataset.  Arrows on the left show how the Reservoir  states are updated and feed into the transformer-based models as inputs. }

  \label{fig:reservoir_transformer_example}
  \vspace{-1.2em}
\end{figure}

\paragraph{ResFormer Architecture}

The core innovation of our ResFormer lies in the integration of Reservoir and Transformer models.
Figure~\ref{fig:reservoir_transformer} depicts the architecture of our ResFormer, which consists of two cascaded modules tailored for different context lengths. 

For longer contexts, an Reservoir models the entire context relevant to the current utterance, drawing from all available training corpora. For example, in a script containing multiple seasons and episodes, the Reservoir learns the emotion of a character based on the current sentence within the larger context throughout the TV series, i.e., scripts of all seasons and episodes.  In scenarios where the corpus consists of several unrelated books, each book is learned in its entirety, with the Reservoir re-initialized randomly at the beginning of each book's first sentence. 

For shorter contexts, a Transformer-based module captures token-level dependencies within individual sentences. Each sentence is initially processed by the Transformer-based model to encode within-sentence dependencies, generating sentence embeddings subsequently combined with outputs from the Reservoir module. The Reservoir reads sentence after sentence in their embedded forms and then update the Reservoir states on all the context of a corpus without training. 

This combined representation is fed into a neural network block. The neural network block can be of any architecture, and in this paper, we showcase on the Transformer model. We apply the CLS token's hidden state to incorporate the sentence level information, enabling comprehensive contextual dependency modeling. 
This cascaded approach effectively models both long-term dependencies and conventional token-level relationships.

Our proposed reservoir method enables the Transformer architecture to process an unlimited number of input tokens. Our primary contributions are as follows:

\begin{enumerate}
\item We propose the ResFormer architecture, capable of theoretically handling arbitrarily long inputs with linear time complexity.
\item We develop a cascaded learning framework to effectively manage long and short contextual information separately.
\item We introduce a cross attention readout mechanism leveraging cross-attention, significantly enhancing the performance of Reservoir.
%\item We implement a group Reservoir approach to further improve the effectiveness of ResFormer.
\item We empirically validate our method, demonstrating notable performance improvements on emotion and intent detection tasks.
\end{enumerate}
%\end{enumerate}

Our experiments demonstrate that our ResFormer significantly enhances performance on NLP classification tasks. Specifically, we observe accuracy increase of +19.9\% over ModernBERT~\cite{warner2024smarterbetterfasterlonger} and +22.3\% over DeepSeek-Qwen-1.5B~\cite{deepseekai2025deepseekr1incentivizingreasoningcapability} on the EmoryNLP dataset, and up to +8.58\%, +8.00\%, and +14.6\% over ModernBERT~\cite{warner2024smarterbetterfasterlonger}, DeepSeek~\cite{deepseekai2025deepseekr1incentivizingreasoningcapability}, and Longformer~\cite{beltagy2020longformer} on the MELD~\cite{poria2019meldmultimodalmultipartydataset} dataset, respectively, in addition to the consistent improvements on MultiWOZ 2.2~\cite{zang-etal-2020-multiwoz} and IEMOCAP~\cite{busso2008iemocap}.

\section{Problem Definition}

A corpus, such as a book, a series of conversations, and all episodes and seasons of a TV-series, can be represented in the text form of a sequence of sentences $\mathbf{u}_1^I=\mathbf{u}_1, \mathbf{u}_2 \cdots \mathbf{u}_i \cdots \mathbf{u}_I$ $(i\in 1, 2, \cdots, I)$, here I is the total number of sentences in the corpus and each individual sentence $\mathbf{u_i}$ is defined as $\mathbf{u_i}=w_1 \cdots w_j \cdots w_J$, where $J$ is the sentence length. A training dataset can be composed of one or multiple corpora.

In a textual sequence classification task, such as emotion and intent detection, we aim to predict the best class $\hat{c}$ for the sentence $\mathbf{u_i}$
based on all previous relevant sentences from $\mathbf{u_1^{i-1}}$:
\begin{align}\label{eq:def}
\hat{c_i}=\argmax_c \Pr&(c | \mathbf{u}_i; \mathbf{u}_1^{i-1})
\end{align}
In Section~\ref{sec:reservoir_transformer}, we will describe our ResFormer  to infer $\Pr$ and predict the sequence class. 

\section{ResFormer}\label{sec:reservoir_transformer}

\paragraph{Architecture} ResFormer $\mathbb{RT}$ is named after reservoir computing~\cite{gallicchio2017deep} $\mathbb{R}(\cdot)$ and Transformer $\mathbb{T}(\cdot)$ architecture it builds upon. The key innovation of $\mathbb{RT}$ lies in integrating two distinct memory mechanisms to capture information across different levels of context. 
This all-context-aware memory framework does not only tackles the challenge of processing long contexts but also selectively emphasizes the most relevant recent inputs. The two memory modules work in tandem to model dependencies at multiple scales when the context length grows:

\begin{itemize}

\item The long-term memory module (LTM) in Section~\ref{subsec:RC} processes the entire context, allowing it to capture extended dependencies across all previous input sentences in a corpus.

\item The short-term memory module (STM) in Section~\ref{sec:embedding}, on the other hand, focuses on local dependencies within individual sentences.

\end{itemize}

Together, these modules enable the model to efficiently manage and utilize context across both global and local levels. 
Now, we will describe these two modules to handle different ranges of input lengths, i.e.,  LTM realized with reservoir, while STM implemented as Transformer: 

\begin{equation}\label{eq-rt1}
    {\hat{y}}_i = \mathbb{T}( \mathbb{R}(\epsilon(\mathbf{u}_1^{i-1})) \uplus \epsilon({\mathbf{u}_{i}})) 
\end{equation}

Here, $\hat{y}$ is the final output of the class prediction $\hat{c}$ in Equation~\ref{eq:def}. The reservoir $\mathbb{R}(\cdot)$ takes all time inputs sequentially, i.e., the long-term historical context from the initial time step to the latest time step  and outputs reservoir states as a compressed full context representation. 
 The operator $\uplus$ represents a cross attention that combines input embeddings with reservoir state output, and then feeds it into a Transformer model. The resulting effective input size to the Transformer is the same as the baseline input size. $\mathbb{T}(\cdot)$ is a Transformer model that learns from the current sentence $\mathbf{u}_{i}$.
 
Below, we will describe how to model input data embedding $\epsilon$ (Section~\ref{sec:embedding}), reservoir $\mathbb{R}$ (Section~\ref{subsec:RC}),  combination operator $\uplus$ (Section~\ref{sec:combine}), Transformer $ \mathbb{T}$ (Section~\ref{sec:STM}), and training  (Section~\ref{sec:seq_batch}). 

\subsection{Embedding \texorpdfstring{$\epsilon$}{}}\label{sec:embedding}

First, we use the Transformer model to embed the input sentences. For a given input sentence $\mathbf{u}$, each of the $J$ tokens $w_1^J$ are fed to the embedding layer (Equation \ref{eq:stm})
as input and we get the sentence embedding matrix $\epsilon(\mathbf{u})$ as the output. 

\begin{align}
     \epsilon (\mathbf{u})= \epsilon (w_1)\oplus \cdots \oplus  \epsilon (w_J)\oplus(w_{CLS})\label{eq:stm}
\end{align}

Here, $\epsilon(w_j)$ denotes the output of the embedding layer for the token $w_j$, obtained from the final hidden layer of the Transformer in the previous iteration (see STM in Section~\ref{sec:STM}). 

Since our focus is on sequence classification tasks, a special classification token ``CLS'' ($w_{\text{CLS}}$) is incorporated during the embedding process. 
The CLS token has been widely used in sequence classification models such as BERT \cite{devlin2019bert}, ModernBERT \cite{warner2024smarterbetterfasterlonger}, BART \cite{lewis2019bart}, and DeepSeek \cite{deepseekai2025deepseekr1incentivizingreasoningcapability}. The CLS token is prepended to the input tokens (e.g., [CLS], sentence tokens) to capture a sentence-level representation.
At each self-attention layer, it attends to all other tokens, aggregating contextual information. After the final layer, the CLS token's hidden state \(\mathbf{h}_{\text{[CLS]}}^{(L)}\) summarizes the entire input and is fed into a classifier The loss function trains the CLS token to capture task-specific context such as sentiment or intent.

\subsection{LTM \texorpdfstring{$\mathbb{R}$}{} }\label{subsec:RC}

The LTM module processes all input sentences in the corpus and produces reservoir states. 
We use the Leaky Integrator reservoir (LI-reservoir) model \citep{jaeger2007optimization}, a variant of the basic reservoir, in which leaky integrator reservoir units are adopted:
\begin{align}\label{eq:res} \mathbf{x}_{t} &= (1-\alpha)\mathbf{x}_{t-1} + \nonumber\\ &\alpha\tanh(\mathbf{W}_{in}\mathbf{h}_{t} + \boldsymbol{\theta} + \mathbf{W}\mathbf{x}_{t-1}), \end{align}

The parameters are defined as follows:

\begin{itemize}[leftmargin=*]
    \item \textbf{Reservoir State}: The reservoir state vector at time $t$ is denoted as $\mathbf{x}_{t} \in \mathbb{R}^{N_r}$, capturing the network's memory of past inputs. The initial state $\mathbf{x}_0$ is either set to zero or initialized randomly. The weight matrices associated with the reservoir are initialized randomly and remain fixed throughout training. The leaky integration parameter $\alpha \in [0,1]$ and the spectral radius $\rho$ are optimized on the validation set using Powell's algorithm.

    \item \textbf{Activation Function}: The nonlinearity in the reservoir is introduced via the element-wise hyperbolic tangent function, $\tanh(\cdot)$.

    \item \textbf{Input and Output Dimensions}: $N_u$ is the number of input units, and $N_r$ is the number of reservoir (hidden) units.

    \item \textbf{Leaky Integration and Hyperparameters}: The leaky integration parameter $\alpha$ regulates the memory of the reservoir; smaller values favor longer memory. The spectral radius $\rho$ is scaled to satisfy the Echo State Property (ESP)~\citep{jaeger2004harnessing,tivno2007markovian}, and the sparsity level is configured accordingly. These hyperparameters are tuned on the validation set. 

    \item \textbf{Weight Matrices}:
    \begin{itemize}
        \item $\mathbf{W} \in \mathbb{R}^{N_r \times N_r}$ is the fixed, recurrent reservoir weight matrix, generated randomly with elements drawn independently from a Gaussian distribution.
        \item $\mathbf{W}_{in} \in \mathbb{R}^{N_r \times N_u}$ is the input-to-reservoir weight matrix, where each element is sampled uniformly from $[-\sigma_{in}, \sigma_{in}]$, with $\sigma_{in}$ as the input scaling factor.
        \item $\boldsymbol{\theta} \in \mathbb{R}^{N_r}$ is the bias-to-reservoir vector, also initialized from the same distribution.
    \end{itemize}
    These matrices are not updated during training; instead, they remain fixed at their initial random values.
\end{itemize}

\paragraph{Nonlinear Readout:} \label{sec:lr}

In reservoir computing, the readout component is responsible for mapping reservoir states to the output space. Rather than using a conventional linear readout, we employ a single-layer multilayer perceptron (MLP) to compute the output at each time step $t$:

\begin{equation} \label{linear_rd}
\mathbf{h}_t = \sigma \left(\mathbf{W}_{out} \mathbf{x}_{t} +  \boldsymbol{\theta}_{out}\right),
\end{equation}

where $\mathbf{h}_t \in \mathbb{R}^m$ is the readout with dimensionality $m$, $\mathbf{W}_{out} \in \mathbb{R}^{m\times N_r}$ is the reservoir-to-readout weight matrix, and $\boldsymbol{\theta}_{out} \in \mathbb{R}^{m}$ is the bias vector. The activation function $\sigma$ is chosen to be ReLU.

Integrating a nonlinear readout into reservoir computing significantly boosts the model's expressiveness, enabling it to capture complex patterns that linear readouts cannot. This leads to improved performance on tasks with nonlinear input–output relationships, such as classification and sequence modeling. Moreover, using a nonlinear readout improves task performance without the need to increase the size or complexity of the reservoir itself, making it a cost-effective enhancement. It also enables greater flexibility in adapting to different output structures and supports modern training techniques, such as gradient-based optimization, when appropriate. Overall, the nonlinear readout serves as a powerful and efficient extension to the reservoir architecture, yielding improved accuracy and generalization.

\paragraph{Group reservoirs:} \label{sec:gr}

As ensemble reduces the variants of the prediction for a more reliable output, we integrate multiple reservoirs' nonlinear readout layers for the stability of our prediction. We consider \(L\) reservoirs, each with distinct decay rates (\(\alpha\) and \(\rho\)), initialized randomly \citep{gallicchio2017deep} and concatenate  \(\oplus\)  each reservoir's output to get the group reservoirs' output \(\mathbf{o}_t\):
\begin{equation}\label{eq:gr2}
\mathbf{o}_t = \mathbf{h}_t^1 \oplus \mathbf{h}_t^2 \oplus \ldots  \oplus \mathbf{h}_t^L
\end{equation}

Thus, the reservoir computing function is 
\begin{equation}\label{eq:R}
\mathbf{o}_t = \mathbb{R}(\mathbf{h}_{1:t})
\end{equation}

\iffalse
Substituting $\mathbb{R}(\mathbf{h}_{1:t})$ with $\mathbf{o}_t$, Equation~\ref{eq-rt2} is represented as
\begin{equation}
    \mathbf{\hat{u}}_{t+1:t+\tau} = \tilde\epsilon(\mathbb{M}({\mathbf{h}_{t-k+1:t}} \uplus \mathbf{o}_{t}))\label{eq-rt3}
\end{equation}
\fi

\subsection{Combination \texorpdfstring{$\uplus$}{}} \label{sec:combine}

We use a cross-attention block $\uplus$  to combine the group reservoir output $\mathbf{o}_{t}$ from Equation~\ref{eq:R} and the embeddings $\epsilon(\mathbf{u}_{t})$  to feed into STM in Section~\ref{sec:STM}:

\begin{equation}\label{eq:crs-attn}
\begin{split}
\epsilon({\mathbf{u}_{t}}) \uplus \mathbf{o}_{t} &=
\mathcal{F}(\text{softmax}\left(\mathcal{K} \cdot (\mathbf{o}_t W^V)\right)+\mathbf{o}_t)\\
\mathcal{K}&=\frac{(\epsilon({\mathbf{u}_{t}})  W^Q)(\mathbf{o}_t W^K)^T}{\sqrt{d_k}} 
\end{split}
\end{equation}

We use $\mathcal{K}$ to denote the key, query, and value component.  $W^Q, W^K, W^V$ matrices are  learnable, with $d_k$ as an indication of the dimensions of the key. We initialize them randomly and train them together with the STM in Section~\ref{sec:STM}.  Besides, $\mathcal{F}$ is the layer norm of the feedforward neural networks with two layers of linear transformation with 768 neurons, and one layer of ReLU function in between, corresponding to the layer normalization operation and feed forward operation proposed in Transformer paper~\cite{vaswani2017attention}.

Here is an example to elaborate on how the reservoir outputs are paired with the embedded inputs. The reservoir state for time step $1$ is $\mathbb{R}(\mathbf{h}_1)$, and the reservoir state for time step $2$ is updated by reading the second time step values $\mathbf{h}_2$ into $\mathbb{R}(\mathbf{h}_1)$ to generate $\mathbb{R}(\mathbf{h}_1^2)$, and the reservoir state for time step $3$ is updated by reading the third time step values $\mathbf{h}_3$ into $\mathbb{R}(\mathbf{h}_1^2)$ to generate $\mathbb{R}(\mathbf{h}_1^3)$. Now, say that we set $k = 2$, we can make the sequence classification $y_3$ based on the cross attention between $\mathbb{R}(\mathbf{h}_1^3)$ (with an output size of $k=2$) and $\mathbf{h}_2^3$ shown in Equation \ref{eq:crs-attn}. For the next time step, we read the $h_4$ value into the $\mathbb{R}(\mathbf{h}_1^3)$ and generate $\mathbb{R}(\mathbf{h}_1^4)$. By using the cross attention between $\mathbb{R}(\mathbf{h}_1^4)$ and $\mathbf{h}_3^4$, we can make the latest sequence classification $y_4$.

\subsection{STM}\label{sec:STM} 

We use Transformer~\cite{vaswani2017attention} to implement STM.  
Note, that the STM's only inputs are embeddings, initialized using ModernBERT~\cite{warner2024smarterbetterfasterlonger} with positional embeddings. 
After that,  the LTM output states and the concatenated embeddings are 
fed into the STM to perform the prediction task. Equation~\ref{eq-rt1} is defined as: 
\begin{align}
 \langle y_{t}, \mathbf{h}_{t} \rangle &=\mathbb{T}(  \mathbf{o}_{t} \uplus \epsilon(\mathbf{u}_{t}) )\label{eq:neural-net}   
 \vspace{-1em}
\end{align}
Here, $\mathbf{o}_t$ is the LTM output states from Equation \ref{eq:crs-attn},  $y_{t}$ is the predicted sequence class, $\mathbf{h}_t$ is the final layer output of the model $\mathbb{T}$, and $\mathbb{T}$ is the Transformer model.  \textit{The STM is model-agnostic and can be implemented using any neural network architecture}, such as RNNs, CNNs, LSTMs, and can be applied within encoder-only models (e.g., BERT~\cite{devlin2018bert}, ModernBERT~\cite{warner2024smarterbetterfasterlonger}), decoder-only models (e.g., DeepSeek~\cite{deepseekai2025deepseekr1incentivizingreasoningcapability}), or encoder-decoder architectures (e.g., T5~\cite{raffel2023exploringlimitstransferlearning}).

\subsection{Training and Parallelization}\label{sec:seq_batch}

\paragraph{Training Loss}
Given $\tilde{y}_t$ as the true label and $\hat{y}_t$ as the prediction at iteration $t$, we use the cross-entropy loss function as our objective:
\begin{align}
\mathcal{L}(\tilde{y}, \hat{y}) = -\frac{1}{T} \sum_{t=1}^{T} \tilde{y}_t \log(\hat{y}_t)
\label{eq:cle_loss}
\end{align}

\paragraph{Batch Parallelization}

Training models with an integrated reservoir requires processing the dataset sequentially to capture the inherent memory dependencies between samples, as the computation for each input depends on the reservoir state produced by its predecessor. This sequential nature makes traditional batch training and standard Transformer parallelization impractical. To address this, we propose a novel parallelization method tailored for our ResFormer architecture, which processes input \textit{sentence by sentence}, while maintaining the necessary reservoir state transitions. The method proceeds in the following steps:

\paragraph{Batch Parallelization}

To support efficient training while preserving the sequential memory dynamics of the reservoir, we introduce a custom batch-parallelization strategy, as follows:

\begin{enumerate}[leftmargin=*]
\item At the start of a new corpus, the reservoir state is randomly initialized and trained from scratch to learn the long-term dependencies inherent in the sequence.

    \item Multiple sentences are grouped into a batch (e.g., batch size of $4$). Each sentence is independently embedded using the ModernBERT embedding layers~\cite{warner2024smarterbetterfasterlonger}.

    \item Each embedding is combined with the current reservoir state using the $\uplus$ operation, as described in Section~\ref{sec:combine}. All sentences in the batch use the same input reservoir state, specifically, the final state from the previous batch.

    \item These combined representations are processed in parallel by the STM, producing a hidden state for each sentence.

    \item The resulting hidden states are then passed \textit{sequentially}, in sentence order, through the reservoir to update its internal state. This ensures that temporal dependencies across sentences are preserved.

    \item The final reservoir state, after processing the entire batch, is carried forward and used as input for the next batch.

\end{enumerate}

\begin{table}[!t]
\centering
\caption{Comparison of time and memory complexity. }
\small
\resizebox{0.5\textwidth}{!}{
\begin{tabular}{l|c|c}
    \hline
    \textbf{Model}            & \textbf{Time } & \textbf{Memory }  \\  \hline
    Transformer   & $O(K^2d)$  & $O(Kd+K^2)$\\ 
    RNN      & $O(Kd^2)$  & $O(Kd)$\\  
   Longformer & $O(Kd^2+gKd)$ & $O(Kqd)$\\
    %BigBird & $O(Kd^2 +Kxd+gKd)$ & \\
    Mamba & $O(Krd)$ & $O(rd^2)$\\\hline
    RT &  $O(Kqd)$   &   $O((qd+q^2+n^2)/B)$ \\ 
    RT(Batch)& $O(Kqd)$& $O(qd+q^2+n^2)$ \\\hline
\end{tabular}}
\vspace{-1em}
\label{tab:mem_cmp}
\end{table}
%\Jia{Table complexity: q?}

\paragraph{Complexity: }
Table~\ref{tab:mem_cmp} shows the time and memory comparison of our method with other popular models. Here, $K$ denotes the input sequence length, $q$ is the sentence length (or window size in the case of \textsc{Longformer~\cite{beltagy2020longformer}}), $g$ refers to the number of tokens used for global attention, %$r$ is the rank in the low-rank projection of the state space for the \textsc{Mamba} model~\cite{gu2023mamba}, 
$d$ is the hidden dimension for each model, $B$ represents the batch size, and $n$ is the number of neurons the reservoir. Transformer has a time complexity of $O(K^2 \times d)$, which becomes impractical to compute as the input length $K$ grows large due to the quadratic complexity. 

In contrast, our ResFormer separates processing into two components: Short-Term Memory (STM) and Long-Term Memory (LTM). The LTM module operates with linear time complexity $O(K)$ relative to input length, enabling efficient handling of long contexts. While the STM retains the standard quadratic complexity of attention mechanisms, but its input length can be fixed to a constant (even a large one), allowing the overall model to scale effectively to extremely long sequences without incurring quadratic cost across the full context.

\section{Experiments and Results}
\label{sec:experiments}

In this section, we present experiments demonstrating that ResFormer effectively handles contexts of arbitrary length while fully leveraging long-range dependencies.

\subsection{Experimental Setup}

\paragraph{Data and pre-processing: }
We present the results on four sequence classification datasets: Multimodal EmotionLines Dataset (MELD)~\cite{poria2019meldmultimodalmultipartydataset}, Multi-Domain Wizard-of-Oz (MultiWOZ 2.2)~\cite{zang-etal-2020-multiwoz}, EmoryNLP~\cite{zahiri2017emotiondetectiontvtranscripts}, and the IEMOCAP multimodal dataset~\cite{busso2008iemocap}. Each dataset is split into training, validation, and test sets following the setup in~\citet{park2022efficient}. 

All datasets used are multimodal (text and visual/audio), but we focus solely on textual data (e.g., utterances) for training and evaluation, such as in MELD and IEMOCAP, to isolate the $\mathbb{RT}$ model’s language understanding capabilities. While our approach is text-based, the model's architecture remains flexible and could be extended to multimodal inputs in future work.

\paragraph{Baselines: }
We compare our method with several Transformer-based baselines, including \textsc{Longformer}~\cite{beltagy2020longformer}, \textsc{ModernBERT}~\cite{warner2024smarterbetterfasterlonger}, and \textsc{DeepSeek-Qwen-1.5B}~\cite{deepseekai2025deepseekr1incentivizingreasoningcapability}, all fine-tuned with \textsc{LoRA}~\cite{hu2021loralowrankadaptationlarge}. For all models, we use a weight decay of $0.01$, a learning rate of $0.0002$, a batch size of $16$, and set the \textsc{LoRA}~\cite{hu2021loralowrankadaptationlarge}  $\alpha$ parameter to $8$.

\paragraph{Model Training: }
\textsc{ModernBERT}~\cite{warner2024smarterbetterfasterlonger} serves as the Transformer backbone in our model. We set the attention dropout to $0.1$, weight decay to $0.01$, and learning rate to $0.0002$. 
For the LTM component, we utilize five distinct reservoirs, each initialized with different hyperparameter configurations. Specifically, the reservoir sizes range from 1500 to 1900, spectral radii from $0.7$ to $0.9$, leaky integration values from $0.48$ to $0.52$, and sparsity levels from $0.4$ to $0.6$. These values are optimized on the validation set using the power iteration method.

\begin{table}[!t]
\begin{center}
\caption{Intent Detection on Multioz 2.2.; Emotion classification on EmoryNLP. DSLora: DeepSeek1.5B+Lora. Mem: Memory in GB. Time: Training time in hours.}\label{tab:baseline}
\scalebox{0.86}{
\begin{tabular}
{c|c|c|c|c}\hline
\textbf{Model} & \textbf{Accuracy} & \textbf{F1} & \textbf{Mem}  &\textbf{Time} \\\hline
\multicolumn{5}{c}{Intent Detection (MultiWOZ 2.2)}\\\hline
DSLora &0.822&0.724&30&32 \\
ModernBERT &0.841&0.757& 15&22 \\
RT&0.840&0.763&8&46 \\\hline
\multicolumn{5}{c}{Emotion Classification (EmoryNLP)}\\\hline
DSLora &0.310&0.256&15&3 \\
ModernBERT &0.316&0.252& 10&3 \\
RT&0.379&0.328&3.6&6 \\\hline
\multicolumn{5}{c}{Emotion Detection (MELD)}\\\hline
DSLora &0.513&0.364&22&8.37 \\
ModernBERT &0.516&0.377& 13& 7.19\\
Longformer& 0.482&0.355& 2.5&2 \\
LSTM alone & 0.326 &0.287& 2.1& 1.33 \\
LSTM+Reservoir & 0.394&0.311& 0.81& 2.23\\
RT&0.557&0.529&4.4&10 \\\hline 
\multicolumn{5}{c}{Emotion Detection (IEMOCAP)}\\\hline
DSLora &0.411&0.425&5.6&3.9 \\
ModernBERT &0.413&0.431& 5.6&4 \\
RT&0.437&0.457&1.87&8 \\
RT (Batch)& 0.431&0.442& 5.4& 4\\\hline 
\end{tabular}
}
\end{center}
\vspace{-1em}
\end{table}

\paragraph{ Results: }
Table~\ref{tab:baseline} shows the accuracy (in percentage) of our ResFormer compared to several baseline models. Across all three tasks, our model consistently achieves strong performance.

\begin{itemize}[leftmargin=*]

\item \textbf{Intent Detection:}  
On the MultiWOZ dataset, ResFormer performs comparably to \textsc{ModernBERT}~\cite{warner2024smarterbetterfasterlonger}. While it requires slightly longer training time, it is significantly more memory-efficient—consuming only about one-third of the RAM used by both \textsc{ModernBERT} and \textsc{DeepSeek-Qwen-1.5B}~\cite{deepseekai2025deepseekr1incentivizingreasoningcapability}.

\item \textbf{Emotion Classification:}  
On EmoryNLP~\cite{zahiri2017emotiondetectiontvtranscripts}, ResFormer outperforms all baselines by up to +6\% in prediction accuracy. As with the Intent Detection task, it maintains a much lower memory footprint, requiring only one-third of the RAM used by the other models.

\item \textbf{MELD and IEMOCAP:}  
On MELD~\cite{poria2019meldmultimodalmultipartydataset}, ResFormer achieves approximately a +4\% improvement in accuracy over baselines. On IEMOCAP~\cite{busso2008iemocap}, it improves accuracy by +2.6\% over \textsc{DeepSeek} and +2.4\% over \textsc{ModernBERT}.
\end{itemize}

Despite its relatively longer training time, ResFormer delivers substantial memory savings, using only about one-third of the RAM compared to other models. Furthermore, as detailed in Section~\ref{sec:seq_batch}, our custom batching strategy significantly reduces training time, bringing it in line with baseline models while still achieving superior accuracy.

\paragraph{Compatibility with Other Backbones:}
The ResFormer framework is architecture-agnostic and can be integrated with non-Transformer models such as LSTMs. For example, on the MELD dataset, incorporating the reservoir into a bidirectional LSTM increases accuracy from 32.61\% to 39.4\% (Table~\ref{tab:baseline}). This highlights the reservoir's effectiveness in enhancing temporal modeling across a variety of backbone architectures.

\subsection{Ablation Study}\label{sec:abl}

\begin{figure}[!t]
  \centering
  \includegraphics[width=0.9\linewidth]{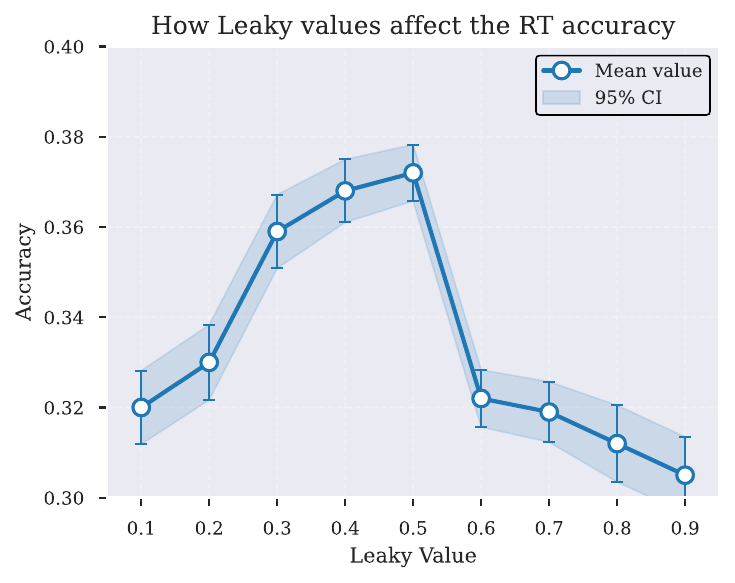}

    \caption{Leaky Values vs. model performance.}

  \label{fig:leaky}
  \vspace{-1.2em}
\end{figure}

\paragraph{Leaky Parameter:}
We investigate how different leaky parameter values ($\alpha$) in the reservoir affect model performance. The leaky parameter controls how much past information is retained in the reservoir state. Figure~\ref{fig:leaky} shows the performance with different $\alpha$ values ranging from 0.3 to 0.7 on the EmoryNLP dataset. We observe that $\alpha = 0.4\sim0.5$ achieves the best performance, suggesting that moderate memory retention works better than either very short or long memory spans.

\begin{figure}[!t]
  \centering
  \includegraphics[width=.9\linewidth]{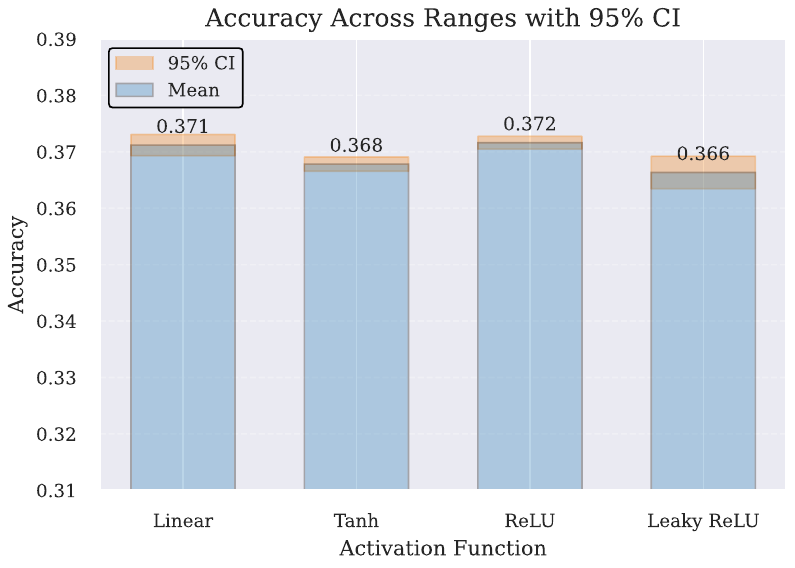}

    \caption{Activation Function vs. model performance.}

  \label{fig:readout}
  %\vspace{-1.2em}
\end{figure}

\paragraph{Nonlinear Readout:}
Figure~\ref{fig:readout} compares the performance of various activation functions in the reservoir readout layer on the EmoryNLP dataset. The tested activations include Linear, Tanh, ReLU, and Leaky ReLU. While ReLU consistently achieves the best results, the overall improvement from using nonlinear activations is modest.

\begin{figure}[!t]
  \centering
  \includegraphics[width=.9\linewidth]{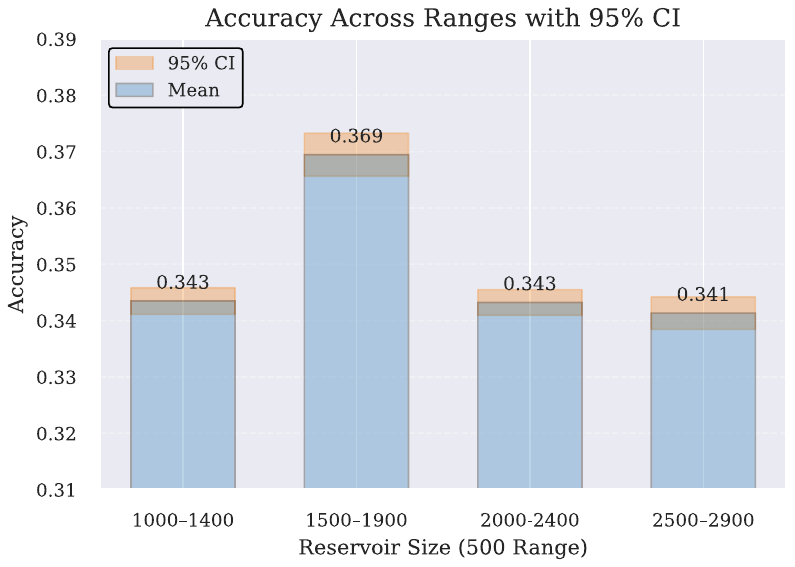}

    \caption{Reservoir size vs. model performance.}

  \label{fig:reservoirsize}
  %\vspace{-1.2em}
\end{figure}

\paragraph{Reservoir Size}
We investigate the impact of reservoir size on model performance, as it determines the dimensionality of the internal state and its capacity to capture temporal dependencies. Figure~\ref{fig:reservoirsize} shows results on EmoryNLP for reservoir sizes ranging from 1000 to 3000 neurons. To effectively analyze this, we use five reservoirs per model, each differing by 100 neurons (e.g., 1000, 1100, 1200, 1300, 1400), allowing us to identify which size range yields the highest accuracy.

\iffalse
\begin{table}[!ht]
    \centering
    \resizebox{0.3\textwidth}{!}{ 
    \begin{tabular}{|c|c|} 
        \hline
        \textbf{Reservoir Size ($N$)} & \textbf{Accuracy} \\ 
        \hline
        1,000--1,500 & 0.342 \\ 
        1,500--2,000 & 0.373 \\ 
        2,000--2,500 & 0.345 \\
        2,500--3,000 & 0.343 \\
        \hline
    \end{tabular}}
    \caption{Impact of different reservoir sizes on model performance using EmoryNLP dataset. Higher accuracy indicates better performance (after 3 epochs).} 
    \label{tab:reservoir_size}
    \vspace{-0.5em} 
\end{table}
\fi

\begin{table}[t!]
\centering
\caption{Hyperparameters for each reservoir, identified by its ID.}
\scalebox{.86}{
\begin{tabular}{ccccc}

ID & Reservoir  & Spectral  & Leaky  & Sparsity \\
ID &  Size &  Radius &  Values &  \\\hline
 \#1 & 1500 & 0.9 & 0.48 & 0.6 \\\hline
 \#2 & 1600 & 0.85 & 0.49 & 0.55 \\\hline
 \#3 & 1700 & 0.8 & 0.50 & 0.5 \\\hline
 \#4 & 1800 & 0.75 & 0.51 & 0.45 \\\hline
 \#5 & 1900 & 0.7 & 0.52 & 0.4 \\

\end{tabular}}
\label{tab:reservoir_params}
\end{table}

Table~\ref{tab:reservoir_params} details the reservoir sizes within the group reservoir and their corresponding hyperparameters.

\paragraph{Combination Method Comparison:}
Ablation on EmoryNLP (Table~\ref{tab:combination}) shows that replacing our cross-attention with simple concatenation or element-wise addition drops accuracy drastically (17.3\% and 14.6\% vs. 37.9\%). This confirms cross-attention as a crucial design choice for effectively integrating reservoir states with Transformer layers, validating its role in enhancing model performance.

\begin{table}[!t]
\centering
\caption{Combination methods between reservoir states and Transformer layers on EmoryNLP dataset.}
\label{tab:combination}
\scalebox{.86}{
\begin{tabular}{l|c}
\hline
\textbf{Combination Method} & \textbf{Accuracy} \\\hline
Simple Concatenation & 0.173 \\\hline
Element-wise Addition & 0.146 \\\hline
Cross-Attention (Ours) & \textbf{0.379} \\
\hline
\end{tabular}}
\end{table}

\paragraph{Out-of-Domain Generalization:}
To evaluate ResFormer’s robustness beyond its training domain, we conduct cross-dataset tests by training on EmoryNLP and evaluating on MELD. Due to different emotion labels, we manually aligned semantically similar categories (e.g., \textit{Fear}/\textit{Scared}, \textit{Joy}/\textit{Joyful}) to enable fair comparison. As Table~\ref{tab:crossdomain} shows, ResFormer outperforms ModernBERT in this setting (28.6\% vs. 24.9\%). These results highlight ResFormer’s potential for zero- or few-shot learning and suggest that the reservoir-based design supports cross-domain generalization.

\begin{table}[!t]
\centering
\caption{Out-of-domain evaluation results.}
\label{tab:crossdomain}
\scalebox{.86}{
\begin{tabular}{l|c|c|c}
\hline
\textbf{Method} & \textbf{Training} & \textbf{Test} & \textbf{Accuracy} \\\hline
ModernBERT & EmoryNLP & MELD & 0.249 \\\hline
ResFormer & EmoryNLP & MELD & \textbf{0.286} \\\hline
\end{tabular}}
\end{table}

\paragraph{ResFormer Stability:}
We test ResFormer with different random seeds and spectral radii across multiple datasets. As shown in Table~\ref{tab:variance}, performance variance remains consistently low (e.g., 0.00025 on EmoryNLP), indicating stable training and robustness to initialization.

\begin{table}[!t]
\centering
\caption{Performance variance of ResFormer across datasets with different random seeds.}
\label{tab:variance}
\scalebox{.86}{
\begin{tabular}{l|c}
\hline
 \textbf{Dataset} & \textbf{Variance} \\\hline
EmoryNLP & 0.00025 \\\hline
 MELD & 0.00046 \\\hline
 IEMOCAP & 0.00057 \\\hline
 Multi\_OZ & 0.00031 \\\hline
\end{tabular}}
\end{table}

\paragraph{Training Efficiency in No-Batch Setting:} 
We also evaluate ResFormer’s efficiency on MELD by training RT and all baseline models to a setting accuracy level with a batch size of 1, demonstrating that it outperforms both ModernBERT and DeepSeek in training time and memory usage. ResFormer reaches the setting accuracy in 4 hours, compared to 7 and 8.75 hours respectively, while using similar memory to ModernBERT (4.4GB) and less than DeepSeek (5.6GB). Although ResFormer is slightly slower per sample, it converges faster, leading to overall improved training efficiency (Table~\ref{tab:efficiency}).

\begin{table}[!t]
\centering
\caption{Training efficiency and memory usage comparison on MELD (batch size = 1).}
\label{tab:efficiency}
\scalebox{.86}{
\begin{tabular}{l|c|c|c}
\hline
\textbf{Method} & ResFormer & ModernBERT & DeepSeek \\\hline
Time & 4 hrs & 7 hrs & 8.75 hrs \\\hline
Memory  & 4.4 GB & 4.4 GB & 5.6 GB \\
\hline
\end{tabular}}
\end{table}

\section{Related Work} \label{related_work}

Extensive research has focused on efficient long-sequence modeling, including but not limited to the works of~\cite{kitaev2020reformer, kim2020length, beltagy2020longformer, choromanski2020rethinking, katharopoulos2020transformers, zhou2021informer, guo2021longt5, ma2021luna, hua2022transformer, tay2022efficient, bertsch2023unlimiformer, liu2023ring, li2023functional, mohtashami2023landmark, ainslie2023colt5, bulatov2023scaling, martins2021infty, liu2024blockwise, munkhdalai2024leave, tworkowski2024focused, bertsch2024unlimiformer, han2023lm, mohtashami2024random}.

Scaling model size, as in \textsc{LLaMA~3}~\cite{llama3}, is a simple fix but fails for arbitrarily long contexts~\cite{mohtashami2024random, kryscinski2021booksum}. Alternatives include fixed-size context encoding~\cite{kanerva1988sparse}, modified attention to highlight salient tokens~\cite{beltagy2020longformer, zaheer2020big, liu2024blockwise}, heuristic attention~\cite{liu2024blockwise, zaheer2020big}, and sequence compression~\cite{peters2018deep, devlin2018bert}, which risk information loss~\cite{li2024streamingdialogue}. Targeted memory strategies scale beyond one million tokens~\cite{munkhdalai2024leave}, while others offload cross-attention to external memories like $k$-NN~\cite{bertsch2024unlimiformer} or retrieval-augmented attention~\cite{tworkowski2024focused}.

Architectural innovations restructure attention via block-wise~\cite{liu2024blockwise}, ring~\cite{liu2023ring}, or sparse mechanisms~\cite{zaheer2020big}, or embed recurrence in deep models~\cite{munkhdalai2019metalearned, feng2024attention, bulatov2022recurrent, wang2019r, kim2018recurrent, bulatov2023scaling}. State-space models like Mamba~\cite{gu2023mamba} maintain compact representations but still face compression bottlenecks~\cite{li2024streamingdialogue}.

Notably, reservoir computing~\cite{jaeger2001echo, maass2002real, xia2023reservoir} has been integrated with deep architectures for temporal tasks~\cite{wang2023continual}. While promising in speech~\cite{nako2023reservoir, ibrahim2021speech} and time series~\cite{shahi2022prediction, bianchi2020reservoir, platt2022systematic, shen2020reservoir}, it remains underexplored for textual data.

In this work, we introduce a novel framework that captures dependencies across short, medium, and long contexts using fixed-length representations, enabling efficient long-sequence modeling while preserving rich textual information.

\section{Conclusion}

We propose an ResFormer model that efficiently learns long input sequences. The core novelty of our approach lies in the integration of long-term and short-term memory on the model level, which enables learn extremely long context in linear time and a constant memory. This design allows the model to capture temporal dependencies across sequences, thereby significantly improving performance on downstream tasks. %Our method significantly enhances performance over \textsc{ModernBERT} and \textsc{DeepSeek-Qwen} on emotion and intent detection tasks.

\section*{Acknowledgement}
We gratefully acknowledge partial support from the eBay eRUPT Program and ACC-New Jersey (Contract No. W15QKN-18-D-0040), whose invaluable support has been essential to this work. We also thank Abdul Rafae Khan for his initial efforts in this line of research.

\section{Limitations}
ResFormer offers limited benefits for short sequences or tasks with minimal long-range dependencies. While full-attention Transformers may outperform it under unlimited resources, ResFormer excels in realistic, resource-constrained settings by efficiently handling long inputs.

\section{Ethical Considerations}
We use AI to enhance grammar, conducting experiments exclusively on public, non-sensitive datasets. Our goal is to advance efficient and accurate NLP while promoting secure and responsible model.

%\bibliography{latex/Bibs/main,latex/Bibs/custom,latex/Bibs/example_paper,latex/Bibs/output}
%\bibliographystyle{./acl_natbib.bst}

%\newpage
%\input{latex/sections/appendix}

\end{document}